\documentclass[journal]{IEEEtran}

\usepackage{cite}
\usepackage{graphicx}
\graphicspath{{photo/}{generated/phase1_figure3/}}
\DeclareGraphicsExtensions{.pdf,.png,.jpg,.jpeg}
\usepackage{amsmath,amssymb}
\usepackage{booktabs}
\usepackage{tabularx}
\usepackage{array}
\usepackage{multirow}
\usepackage{threeparttable}
\usepackage{url}
\begin{document}

\title{Layered Outer-Loop Control for Disturbance-Robust Multi-Waypoint UAV Arrival}

\author{Runfeng~Ling%
\thanks{Runfeng Ling is with The University of Manchester, Manchester, U.K.}}

\markboth{Layered Outer-Loop UAV Arrival}{Ling: Layered Outer-Loop UAV Arrival}

\maketitle

\begin{abstract}
Disturbance-robust UAV position control is easy to demonstrate in benign simulations but much harder to make fast in approach, well behaved near the target, and credible beyond a single benchmark. This letter presents a layered terminal-control architecture for multi-waypoint UAV position regulation together with a staged evaluation across PyBullet, PX4/Gazebo, and hardware. Phase~I uses a PyBullet benchmark with stochastic wind for rapid structural selection, identifying a controller core that separates smooth approach generation, persistent-bias compensation, and supervised near-target terminal regulation. Phase~II carries only that main architecture into a more demanding PX4/Gazebo closed loop, where the outer-loop controller acts through a cascaded flight stack with delay-sensitive settling and stronger transit-to-hover coupling. This step exposes which benchmark gains survive autopilot-mediated dynamics and which refinements collapse once the loop becomes more deployment-like. In Phase~I, the bare controller attains 0.024~m mean late-stage wind error. In Phase~II, the final controller is selected using a transfer-oriented rule emphasizing absence of benchmark priors, cross-scenario balance, and deployable supervisory logic. Strict is used as the primary reporting reference; the supplementary retrospective Grace analysis shows that part of the residual failure set is sensitive to completion semantics rather than gross waypoint-miss behaviour. The evaluation is completed on one Vicon-tracked Tello stack through a two-level hardware study. Taken together, the results suggest that benchmark success becomes more informative when the main controller design is separated from benchmark-specific refinement and remains defensible under harder closed-loop evaluation.
\end{abstract}


\IEEEpeerreviewmaketitle

\section{Introduction}

\IEEEPARstart{A}{ccurate} UAV position regulation is easy to demonstrate in benign simulations, but much harder to sustain once the task demands fast approach, controlled braking, clean terminal settling, and robustness to disturbance persistence or closed-loop delay. This difficulty is especially visible in multi-waypoint flight, where the controller must repeatedly transition from transit to near-target regulation without turning each waypoint into a small overshoot-and-correction cycle. In this setting, visually smooth trajectories are not enough: many controllers that look convincing during nominal transit degrade once the evaluation emphasizes what happens after apparent arrival, namely late-stage error, rebound, residual oscillation, and disturbance-driven re-separation from the target region.

The problem becomes sharper when the controller is developed inside a benchmark. Target-conditioned residual correction or scenario-specific tuning can improve summary scores within that benchmark without necessarily improving behaviour under a different closed-loop setting. Conversely, compact classical controllers often remain competitive because average-case metrics do not fully expose the terminal-regulation problem. For aerial robotics, the key question is therefore not simply how to obtain a better score in one environment, but how to identify controller structure that remains effective once the evaluation setting becomes more demanding.

This letter addresses that question through a layered terminal-control architecture for multi-waypoint UAV position regulation. The central mechanism is to treat terminal waypoint regulation as a conflict between transit-efficient motion and near-target regulation: command authority that is helpful during approach can produce overshoot, rebound, or residual drift near the waypoint, whereas conservative terminal behaviour can slow the mission and weaken disturbance recovery. The proposed controller resolves this conflict by separating smooth approach authority, persistent-bias compensation, and supervised capture/settle authority within one bounded outer-loop velocity interface. A staged evaluation across PyBullet, PX4/Gazebo, and hardware is then used to test whether this mechanism remains useful beyond the original benchmark.

This framing leads to three connected claims. First, the proposed layered architecture addresses the transit--terminal conflict through three design separations: approach authority versus terminal authority, persistent-bias compensation versus rebound recovery, and benchmark-specific refinement versus transferable mainline control. Second, the staged evaluation across environments tests whether that mechanism remains effective once the loop becomes more demanding, rather than treating benchmark score as sufficient evidence. Third, the paper closes with a two-level hardware study on one controlled hardware stack: an early hardware comparison checks whether the principal Phase-I method distinctions remain visible on the physical platform, and a final hardware study validates the controller family selected after Phase~II.

The quantitative results support all three parts of this narrative. In Phase~I, the bare controller reaches wind-condition late-stage errors of 0.024~m mean and 0.017~m standard deviation without any target-specific map. In Phase~II, the final controller is selected before hardware deployment using a transfer-oriented rule rather than a single-scenario endpoint score; Strict is used as the primary reporting reference and Grace is used only to sharpen the interpretation through completion-semantics sensitivity. The completed real-platform study then reports the Stage-A comparison outcome together with Stage-B mission and terminal-accuracy results, closing the transfer story on one controlled physical platform.

The contributions of the paper are threefold:
\begin{enumerate}
\item We present a layered terminal-control architecture for disturbance-robust UAV position regulation that addresses the conflict between fast transit and clean terminal regulation through smooth approach fade-out, persistent-bias compensation, and supervised capture/settle authority.
\item We organize controller assessment through a two-stage cross-environment evaluation process that distinguishes the main controller mechanism from benchmark-specific refinements and clarifies which design choices remain effective beyond the original benchmark.
\item We provide a two-level hardware study on one Vicon-tracked Tello hardware stack, in which an early hardware comparison checks whether the main Phase-I method distinctions survive on the platform and a final hardware study validates the selected controller family under a shared protocol.
\end{enumerate}

\begin{figure*}[!t]
\centering
\includegraphics[width=\textwidth]{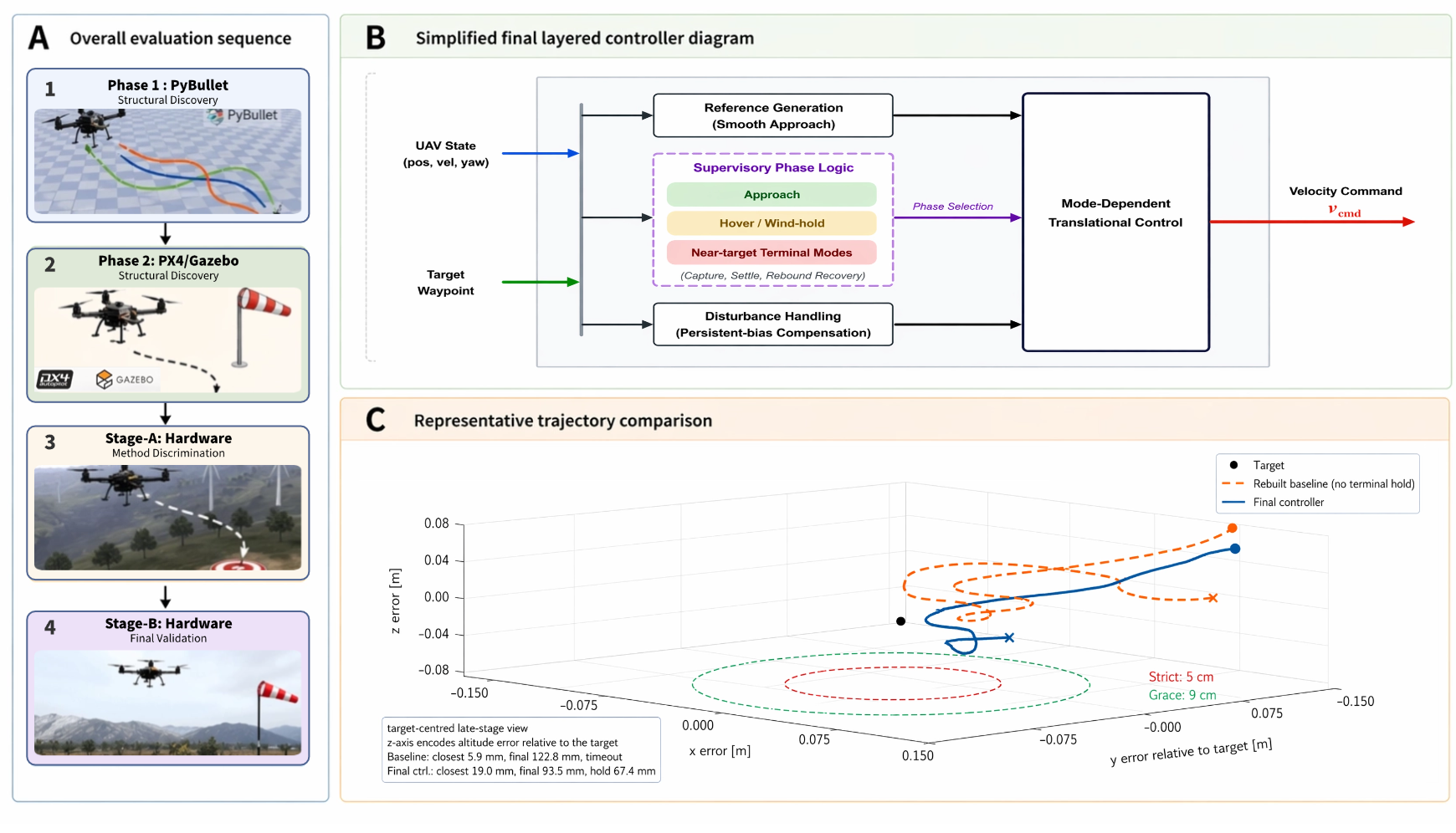}
\caption{Overview of the proposed evaluation sequence and controller family. The study first identifies an effective layered outer-loop structure in a PyBullet benchmark environment, then tests and refines only the main controller architecture in a more demanding PX4/Gazebo closed-loop environment, and finally closes the argument through a two-level hardware study: an early hardware comparison for method discrimination followed by final hardware validation of the selected controller family.}
\label{fig:overview}
\end{figure*}

\section{Related Work and Positioning}

Outer-loop UAV position control has been studied through a broad range of trajectory-tracking and waypoint-reaching approaches, including PID-style feedback, nonlinear and geometric tracking control, optimal-control-inspired formulations, and smoother reference-generation strategies for aggressive but well-behaved motion~\cite{Mahony2012,Bouabdallah2005,Lee2010,Lee2013,Mellinger2011,Mellinger2012,Richter2016,Mueller2015,Faessler2018}. These methods are effective for nominal target reaching and often provide strong average-case performance in simulation. However, their terminal behaviour can remain sensitive to disturbance persistence, transition timing, and the interaction between approach dynamics and near-target stabilization. In particular, controllers that look similar during transit may differ substantially once the task is judged by late-stage settling quality rather than by gross waypoint attainment alone.

Wind rejection and disturbance compensation have likewise been addressed through several established mechanisms, including integral action, disturbance observers, bias adaptation, and disturbance-aware feedforward or rejection loops~\cite{Cabecinhas2014,Chen2016,Han2009,Li2014,Besnard2012}. Such methods are essential whenever the controlled platform is exposed to slowly varying drift or gust-like perturbations. At the same time, existing disturbance-handling components do not by themselves resolve the full terminal-control problem, because near-target performance depends not only on disturbance estimation accuracy but also on how disturbance compensation interacts with braking, handoff, and mode switching as the vehicle transitions from transit to hold.

Related ideas also appear in hybrid and phase-dependent control designs, where different terminal regimes are handled through explicit switching logic, capture modes, or staged convergence policies~\cite{Liberzon2003,Goebel2012,Branicky1998}. This line of work is highly relevant to aerial robotics, since the control objective changes qualitatively as the vehicle moves from long-range tracking to near-target regulation. The present work is aligned with this perspective, but differs in emphasis: rather than treating terminal heuristics as isolated patches, it organizes them as part of a layered terminal-control architecture with explicit supervisory logic and a clear separation between transferable structure and benchmark-specific refinement.

Finally, cross-platform validation remains a central challenge in aerial robotics. Controllers that perform strongly in a single simulator may rely, intentionally or not, on properties of the evaluation environment that do not survive migration to a harder closed loop or to hardware~\cite{Koenig2004,Meyer2012,Meier2015,Furrer2016,Jakobi1995,Song2021}. Conversely, some controller components retain value across platforms precisely because they encode task structure rather than benchmark-specific exploitation. The present paper positions itself in this gap. It does not claim the first use of disturbance observers, phase-dependent logic, smooth reference shaping, or simulator-to-real evaluation in UAV control. Instead, the novelty lies in organizing these established ideas into a layered terminal-control architecture and examining, through staged cross-environment evidence, which parts remain useful beyond the original benchmark.

\section{Proposed Layered Terminal-Control Architecture}
\label{sec:method}

\subsection{Problem Setup}

We formulate multi-waypoint UAV position regulation as an outer-loop velocity-command problem. At each control step, the controller receives the current vehicle state, consisting of position and attitude, together with the active target waypoint and a timestamp or control interval. It returns body-frame translational velocity commands and a yaw-rate command. The objective is not merely to reach the target waypoint, but to do so with fast approach, controlled braking, accurate terminal convergence, and stable hold under disturbances. Because the outer loop cannot command thrust or attitude directly, it must manage terminal behaviour through structured velocity-command generation and explicit near-target supervision.

The central difficulty is that the control objective changes with operating region. Far from the target, the controller should favour efficient transit and smooth approach shaping. Near the target, the task becomes one of terminal regulation: the controller must dissipate residual motion, reject persistent drift, and recover from gust-driven displacement without entering slow oscillatory corrections. The proposed architecture is built around this regime change. Rather than seeking a single monolithic law that behaves identically everywhere, it separates transit-friendly command generation from near-target terminal authority while preserving a common outer-loop interface.

For analysis, the outer-loop-controlled vehicle is represented by the reduced closed-loop abstraction
\begin{equation}
\dot p = v, \qquad
\tau \dot v = -v + \mathrm{sat}(v_{cmd}) + d(t),
\label{eq:plant}
\end{equation}
where \(p \in \mathbb{R}^3\) and \(v \in \mathbb{R}^3\) denote position and translational velocity, \(\tau > 0\) is an effective first-order response constant induced by the autopilot-mediated inner loop, and \(\mathrm{sat}(\cdot)\) denotes the bounded velocity-command interface. The disturbance term \(d(t)\) aggregates wind, steady bias, and residual unmodeled inner-loop effects, and is assumed bounded as
\begin{equation}
\|d(t)\| \le D_{\max}.
\label{eq:disturbance_bound}
\end{equation}
This model is not intended as a high-fidelity plant-identification result. Rather, it provides a compact outer-loop abstraction that captures the two properties most relevant to the present design: finite command bandwidth and persistent bounded disturbance.

\subsection{Architecture Overview}

The final controller is organized around the mechanism that a single fixed command semantics cannot simultaneously be aggressive in transit and conservative near the waypoint. A high-authority approach law improves reach time but can keep injecting motion after the target region should already be regulated; a uniformly conservative law improves settling but weakens approach speed and recovery from disturbance-driven displacement. The controller therefore uses four coupled layers: a smooth approach generator, a mode-dependent translational-control layer, a disturbance-handling layer, and a supervisory terminal-regulation layer.

Each layer addresses a specific part of this conflict. The approach generator shapes the reference so that the controller does not depend on impulsive setpoint changes and can fade feedforward authority near the target. The translational-control layer converts tracking error and motion estimates into nominal command action. The disturbance layer separates persistent-bias compensation from state-triggered terminal recovery instead of forcing one mechanism to handle every disturbance role. The supervisory layer decides which control semantics should dominate, e.g., transit, near-target capture, rebound recovery, or terminal settling. The architecture therefore preserves a continuous command pathway while allowing the balance between reference tracking, damping, disturbance rejection, and terminal correction to change as the vehicle approaches the target.

\subsection{Smooth Approach and Mode-Dependent Control}

For nontrivial waypoint displacements, the controller first generates a smooth approach reference between the current pose and the active target. Assuming obstacle-free local waypoint transitions, this is implemented through a minimum-jerk-type trajectory with a decaying feedforward velocity component, following the broader practice of smooth polynomial reference generation for quadrotors~\cite{Mellinger2011,Mellinger2012,Richter2016,Mueller2015}. The essential design requirement is that the reference remains helpful during transit but gradually loses authority near the target. This fade-out is critical: if the feedforward term persists too long, it continues to inject forward motion precisely when the controller should already be prioritizing braking, capture, and terminal hold.

Let \(p_t\) denote the active target position, \(p_r(t)\) the smooth approach reference, and \(v_r(t) = \dot p_r(t)\) its feedforward velocity. The feedforward authority is modulated by a distance-dependent fade-out factor \(\alpha(\rho) \in [0,1]\), where \(\rho = \|p_t - p\|\), \(\alpha(\rho)\) is monotone nondecreasing with distance, and \(\alpha(\rho) \to 0\) inside the terminal neighborhood. The commanded reference term is therefore written as
\begin{equation}
v_{ref} = \alpha(\rho)\, v_r.
\label{eq:vref}
\end{equation}

At the command level, the translational controller is expressed as a mode-dependent combination of reference tracking, damping, disturbance compensation, and terminal correction:
\begin{equation}
v_{cmd}^w = v_{ref} + K_p^{(\sigma)} e_p - K_d^{(\sigma)} \hat{v} + v_{dist} + v_{term}^{(\sigma)},
\label{eq:main_control}
\end{equation}
where \(e_p = p_r - p\), \(\hat{v}\) denotes the estimated translational velocity, and \(\sigma\) denotes the active supervisory mode. A compact representation of the low-bandwidth disturbance-compensation state is
\begin{equation}
\dot{\hat d} = -\Lambda_d \hat d + \Gamma_d e_p, \qquad
v_{dist} = -\hat d,
\label{eq:disturbance_state}
\end{equation}
where \(\hat d\) is not interpreted as an exact wind estimate, but as a persistent-bias compensation state that absorbs slowly varying drift and model mismatch at the outer-loop level. The terminal-action term is activated only near the target and may be written as
\begin{equation}
v_{term}^{(\sigma)} =
\chi_C(\sigma)\left(K_C e_t - K_{C,v}\hat v\right)
+
\chi_S(\sigma)\left(K_S e_t - K_{S,v}\hat v\right),
\label{eq:vterm}
\end{equation}
where \(e_t = p_t - p\), and \(\chi_C(\sigma), \chi_S(\sigma) \in \{0,1\}\) select the capture and settle corrections, respectively. During transit, \(\chi_C = \chi_S = 0\), so the command is dominated by reference tracking and damping. Near the target, the feedforward authority is progressively attenuated and the same command channel behaves increasingly like a disturbance-robust regulator. The resulting world-frame command is then rotated into a yaw-aligned body frame before being issued as the outer-loop velocity command.

Yaw is regulated separately but according to the same design philosophy. The yaw channel is driven by wrapped yaw error, damping, and bounded handoff behaviour so that heading transients do not destabilize the translational channel during capture and settling. This separation is important because waypoint completion in the present task depends on both position and yaw, while the translational controller should not have to absorb large heading-induced disturbances during the terminal phase.

\subsection{Disturbance Handling and Supervisory Terminal Logic}

The disturbance-handling layer is intentionally split by control role rather than by nominal frequency. A persistent-bias component addresses slowly varying drift and model mismatch, which are typical of wind-like disturbances and steady residual error. A separate state-triggered terminal-recovery component is reserved for near-target situations in which the vehicle is displaced after nominal arrival by rebound or disturbance-driven re-separation. This split avoids a common failure mode in which a single observer or integral mechanism is expected to solve both long-horizon bias rejection and state-dependent terminal recovery.

The supervisory terminal layer determines when the controller should behave primarily as a tracker and when it should behave primarily as a regulator. Mode transitions are defined through explicit conditions involving distance to target, relative speed, persistence inside the target region, and evidence of rebound or re-separation after an apparent arrival. These transitions are part of the method rather than ad hoc implementation detail: they encode the control semantics of the task. A capture mode, for example, is not simply a local gain increase, but a distinct operating regime in which the controller prioritizes energy dissipation, target re-acquisition, and bounded recovery after disturbance-driven displacement. Likewise, the settle regime is not merely ``low gain near the waypoint,'' but a dedicated terminal state entered only after proximity and relative calm have been re-established. In implementation, these transitions are realized through hysteretic entry and exit conditions together with progressive reweighting of command authority rather than abrupt replacement of the full command law, which helps keep the issued velocity command bounded and practically continuous across regime changes.

To make this logic explicit, define the supervisory mode
\begin{equation}
\sigma \in \{\mathrm{T}, \mathrm{C}, \mathrm{S}\},
\label{eq:modes}
\end{equation}
corresponding to transit, capture, and settle. Let \(\rho = \|p_t - p\|\), let \(s = \|\hat v\|\), and let \(b \in \{0,1\}\) denote a rebound flag triggered after prior entry into the terminal neighborhood. A representative switching law is
\begin{equation}
\sigma =
\begin{cases}
\mathrm{T}, & \rho > r_{\mathrm{in}} \text{ and } b = 0, \\
\mathrm{C}, & r_{\mathrm{s}} < \rho \le r_{\mathrm{in}} \text{ or } b = 1, \\
\mathrm{S}, & \rho \le r_{\mathrm{s}},\ s \le s_{\mathrm{s}},\ T_h \ge T_{\mathrm{s}},
\end{cases}
\label{eq:switching}
\end{equation}
where \(T_h\) is a dwell timer inside the settle candidate region. Exit conditions are defined through hysteretic thresholds
\begin{equation}
\mathrm{S} \rightarrow \mathrm{C} \ \text{if}\ \rho > r_{\mathrm{out}} \ \text{or}\ s > s_{\mathrm{out}},
\qquad
\mathrm{C} \rightarrow \mathrm{T} \ \text{if}\ \rho > r_{\mathrm{far}},
\label{eq:exit_conditions}
\end{equation}
with \(r_{\mathrm{out}} > r_{\mathrm{s}}\) and \(s_{\mathrm{out}} > s_{\mathrm{s}}\). These guards are not presented as a full switched-systems proof, but they formalize two practical design properties used throughout the controller: chattering is reduced by hysteresis and persistence conditions, and authority is transferred between modes through bounded reweighting rather than discontinuous replacement of the full command law.

This supervisory structure is what allows the controller to remain both fast in transit and clean near the target. A single fixed law can often do one of these reasonably well, but tends to compromise the other. The layered architecture instead assigns terminal authority to an explicitly supervised near-target regime while preserving efficient outer-loop motion away from the waypoint.

\subsection{What Is and Is Not Part of the Main Method}

The main method claim concerns the core controller architecture described above: smooth approach generation, mode-dependent translational control, persistent-bias compensation, state-triggered terminal recovery, and supervisory terminal regulation. An optional target-conditioned residual refinement is available for the fixed benchmark target distribution, but it is explicitly excluded from the main controller claim. This distinction is essential for the logic of the paper. It prevents benchmark-specific improvement from being conflated with architectural value, and it makes clear why only the core layered controller is carried into the second stage of cross-environment evaluation and, later, into formal hardware validation.

\begin{figure*}[!t]
\centering
\includegraphics[width=0.9\textwidth]{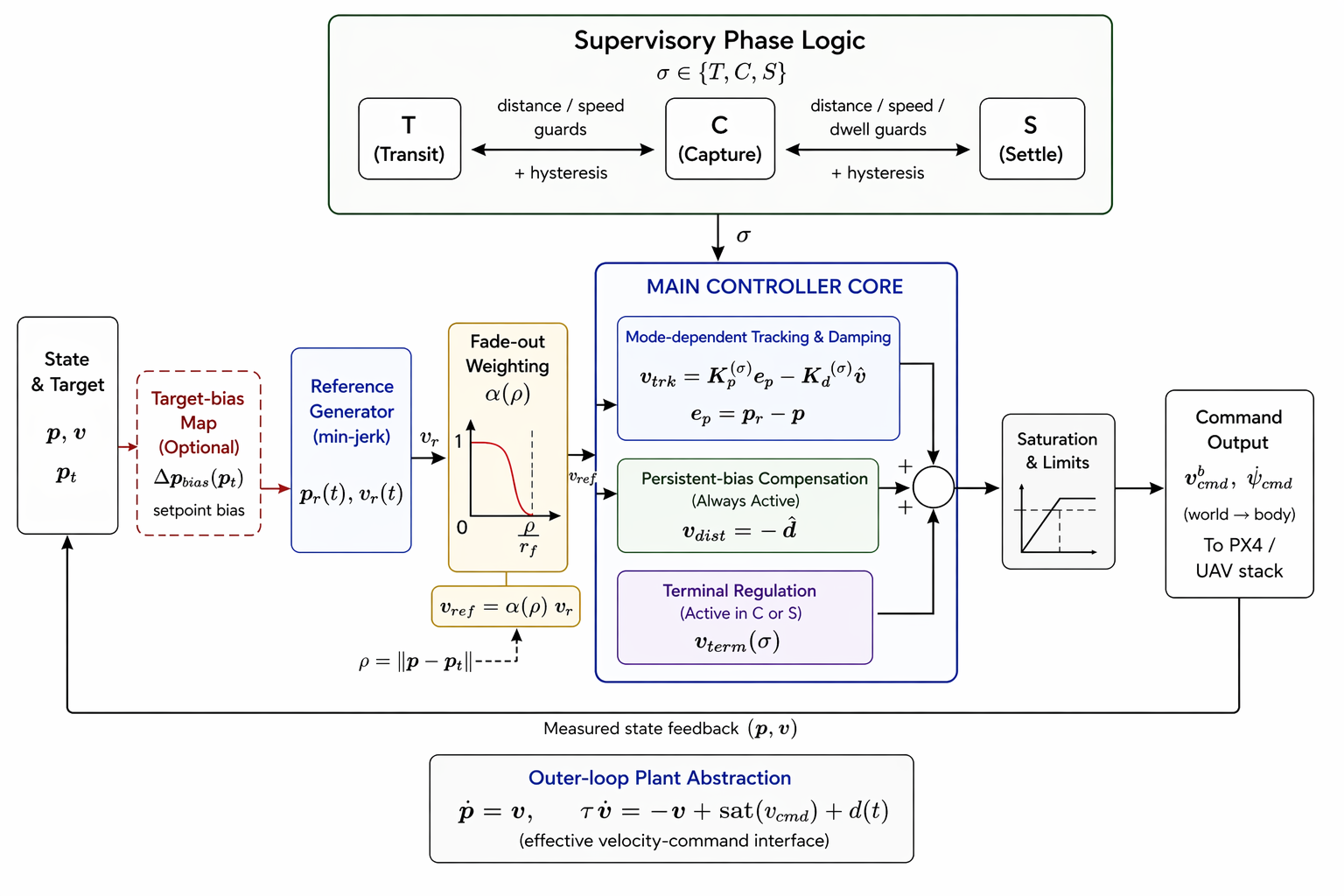}
\caption{Final layered controller architecture. The main controller core separates smooth approach generation, disturbance compensation, and supervised terminal regulation under a reduced outer-loop plant abstraction. A benchmark-specific target-bias refinement is treated as an optional extension rather than part of the main controller claim.}
\label{fig:architecture}
\end{figure*}

\section{Experimental Protocol}

\subsection{Phase I: PyBullet Benchmark Evaluation}

Phase~I uses a PyBullet-based multi-waypoint UAV position-regulation and terminal-stabilization benchmark under stochastic wind disturbances. This phase serves two purposes: controller selection and early structural comparison. PyBullet is used here not simply as a convenient simulator, but as a lightweight and highly repeatable benchmark environment in which different controller structures can be compared quickly across many target and disturbance realizations. The benchmark is evaluated under both no-wind and wind-enabled conditions so that smooth nominal approach and disturbance-robust terminal behaviour can be assessed separately. The main metrics are late-stage position mean, late-stage position standard deviation, mean time-to-reach, and the corresponding yaw statistics. This metric set is intentionally chosen to separate average waypoint pursuit from terminal settling quality. In particular, the benchmark emphasizes what happens after the vehicle has nominally arrived near the target, which is exactly where controller differences become most visible.

Controllers compared in this phase include compact classical baselines, more aggressive fast-approach baselines, the proposed layered controller, and an optional benchmark-specific extension. The purpose of Phase~I is therefore not merely to report the best benchmark score, but to identify which structural elements persist across wind conditions and which improvements depend on benchmark-specific target information.

\subsection{Phase II: PX4/Gazebo Closed-Loop Evaluation}

Phase~II moves to PX4/Gazebo to test whether the proposed terminal-regulation mechanism remains effective under more demanding closed-loop conditions. This phase is not a continuation of benchmark optimization. Compared with the PyBullet benchmark loop, PX4/Gazebo places the outer-loop controller in a cascaded autopilot-mediated closed loop, where command effects are filtered through the flight stack, terminal settling becomes delay-sensitive, and coupling between transit and hold behaviour is more exposed. Accordingly, delay, filtering, and inner-loop coupling are treated as properties of the effective closed-loop plant, and the controller is judged by whether approach fade-out, damping, disturbance handling, and supervisory terminal regulation remain robust under that plant. PX4/Gazebo is therefore used as a harder closed-loop filter for the outer-loop architecture, not as a high-fidelity digital twin of the later Tello hardware platform, consistent with common ROS/Gazebo and PX4-based robotics simulation practice~\cite{Koenig2004,Meyer2012,Meier2015,Furrer2016}. Evaluation focuses on representative difficult scenarios, including combined-disturbance cases and gust-dominated cases, e.g., \texttt{combo\_hard}, \texttt{gust\_step\_y}, and \texttt{steady\_wind\_2ms}.

Only the main controller architecture is carried forward as the main method in this phase. Intermediate controller versions are compared not as isolated final candidates, but as structural checkpoints in the evolution from the PyBullet benchmark to PX4/Gazebo closed-loop evaluation. Before hardware deployment, the carry-forward controller is selected using a transfer-oriented rule: no benchmark- or target-conditioned prior, no scenario-identification requirement at runtime, retained terminal regulation on both representative disturbance classes, and deployable supervisory logic under the same outer-loop interface. For each controller candidate, gains, mode thresholds, and completion thresholds are fixed before batch evaluation and are not retuned per scenario during summary scoring.

The PX4/Gazebo evaluation is executed through a ROS~2 offboard runtime stack rather than through a simulator-internal control hook. The outer-loop controller is wrapped in ROS~2 nodes that subscribe to PX4 vehicle odometry and status topics and publish offboard control mode, trajectory setpoint, and vehicle-command messages through the PX4 ROS~2 interface, with a Micro-XRCE-DDS bridge linking PX4 and ROS~2. Gazebo provides the closed-loop flight environment, while RViz is used for online visualization of targets, reference trajectories, and realized motion. In the current implementation, outer-loop control updates, offboard heartbeat/setpoint publication, and settle/state-supervision logic all run in a fixed 20~Hz timer callback. State delivery is asynchronous: the node subscribes to PX4 odometry and status streams continuously, stores the latest odometry samples in a bounded buffer, and evaluates control from the latest buffered state in the nominal case. In delayed-observation scenarios, the same buffer is queried at a prescribed delay offset so that observation delay is injected explicitly and reproducibly rather than being conflated with ordinary callback timing. Runtime logging is also decoupled from the control loop through an optional reporting period, so experiment traces can be throttled without changing the controller update cadence. This software stack is not part of the claimed control contribution; its role is to provide a reproducible runtime interface in which the same outer-loop logic can be evaluated under a realistic autopilot-mediated execution chain.

\subsection{Evaluation Semantics Sensitivity: Strict vs. Grace}

In addition to scenario difficulty, the evaluation protocol uses Strict and Grace as a supporting lens on terminal behaviour. The Strict criterion is the primary evaluation criterion and represents a harder stationary notion of success, in which the controller is rewarded only when the system satisfies a tightly defined settling condition. The Grace criterion is introduced only as a supplementary, retrospective sensitivity analysis applied uniformly to the retained logs using one fixed threshold set, and represents a task-semantic notion of completion in which the controller is credited once the vehicle has entered and maintained an acceptable terminal tube for a prescribed dwell interval.

This distinction is not introduced as a replacement for the harder criterion, nor as a means of relaxing evaluation post hoc. Rather, it helps interpret the same controller mechanism under two terminal semantics. For outer-loop velocity controllers operating under delay, nonlinear inner-loop effects, and disturbance-driven terminal excursions, different settling criteria can materially alter the apparent ranking of controller variants. Strict is therefore treated as the primary reference criterion, whereas Grace is reported only as a complementary task-semantic sensitivity analysis and is not used to change the controller-selection rule.

For formal evaluation, let \(p(t)\) and \(\psi(t)\) denote the measured position and yaw, let \(p_t\) and \(\psi_t\) denote the active target position and yaw, and let \([T_1,T_2]\) denote a protocol-fixed strict-settling window defined a priori for the waypoint under evaluation. Define
\begin{equation}
\begin{aligned}
e_p(t) &= \|p(t) - p_t\|, \qquad
e_\psi(t) = |\mathrm{wrap}(\psi(t) - \psi_t)|,\\
s(t) &= \|v(t)\|.
\end{aligned}
\label{eq:strict_definitions}
\end{equation}
Under the Strict criterion, a waypoint is counted as settled only if the predeclared strict-settling window \([T_1,T_2]\) satisfies
\begin{equation}
e_p(t) \le \epsilon_p, \qquad
e_\psi(t) \le \epsilon_\psi, \qquad
s(t) \le \epsilon_v,
\qquad \forall t \in [T_1,T_2].
\label{eq:strict_criterion}
\end{equation}
Under the Grace criterion, a waypoint is counted as operationally complete if there exists a time \(T_g\) such that the system enters and remains inside a terminal tube for at least a dwell duration \(T_d\), i.e.,
\begin{equation}
e_p(t) \le R_p, \qquad
e_\psi(t) \le R_\psi,
\qquad \forall t \in [T_g, T_g + T_d].
\label{eq:grace_criterion}
\end{equation}
The key distinction is therefore semantic rather than cosmetic: Strict requires sustained satisfaction of position, yaw, and residual-motion bounds over a protocol-fixed settling window, whereas Grace requires entry into and retention within a practically acceptable terminal region for a prescribed dwell time. The associated thresholds and dwell windows are fixed by the evaluation protocol and held constant for the retained logs used in the Strict/Grace comparison.

\subsection{Real-Platform Validation Protocol}

The real-platform evaluation forms the final validation layer of the paper and is organized in two hardware blocks rather than treated as a single undifferentiated final test. Stage~A is a compact hardware method-comparison block derived from Phase~I. Its role is method discrimination: a small set of representative controllers is executed under the same waypoint logic, safety bounds, and logging protocol in order to test whether the principal structural differences observed in the benchmark stage remain visible on the physical platform. Stage~B then carries forward only the final controller family selected after Phase~II and evaluates it under the finalized multi-waypoint hardware protocol.

Accordingly, the hardware protocol is presented as a completed two-level validation study rather than as a feasibility note. Safety-screening and deployment de-risking flights may be mentioned briefly as pre-validation preparation, but the paper-facing hardware section reports only the finalized Stage-A comparison trials and Stage-B validation trials, together with their summary metrics and representative trajectories or time series. Stage~A is judged primarily by method-discrimination endpoints such as reach rate, reach time, peak overshoot, terminal position or yaw error, and short-window hold motion. Stage~B is judged primarily by final-validation endpoints such as mission completion, total mission time, per-waypoint terminal errors, mean overshoot, and hold-quality metrics. The resulting real-platform section supports hardware-level claims about both early method discrimination and final-controller mission performance on the chosen platform, while still leaving room to discuss the remaining sim-to-real gap and the limits of single-platform validation.

\begin{table*}[!t]
\centering
\caption{Evaluation protocol across the full validation chain. Phase~I uses a lightweight and highly repeatable PyBullet benchmark for controller selection and structural comparison. Phase~II moves the same outer-loop logic into a more demanding PX4/Gazebo closed loop for cross-environment filtering under autopilot-mediated dynamics. The hardware layer closes this evidence chain on one Vicon-tracked Tello stack through Stage-A method discrimination and Stage-B validation of the selected final controller family.}
\label{tab:evaluation_protocol}
\footnotesize
\setlength{\tabcolsep}{4pt}
\renewcommand{\arraystretch}{1.15}
\begin{tabularx}{\textwidth}{
>{\raggedright\arraybackslash}p{1.8cm}
>{\raggedright\arraybackslash}p{2.5cm}
>{\raggedright\arraybackslash}p{3.9cm}
>{\raggedright\arraybackslash}p{4.0cm}
>{\raggedright\arraybackslash}X}
\toprule
\textbf{Stage} & \textbf{Platform} & \textbf{Main Scenarios / Task Structure} & \textbf{Primary Metrics} & \textbf{Purpose} \\
\midrule
Phase~I &
PyBullet benchmark &
Multi-waypoint position-regulation and terminal-stabilization benchmark under no-wind and stochastic-wind conditions &
Late-stage position mean, late-stage position standard deviation, mean time-to-reach, and corresponding yaw statistics &
Rapid controller comparison, early structural filtering, and separation of nominal waypoint pursuit from disturbance-robust terminal behaviour \\

Phase~II &
PX4 SITL + Gazebo + ROS~2 offboard runtime stack &
Representative difficult closed-loop scenarios, including \texttt{combo\_hard}, \texttt{gust\_step\_y}, and \texttt{steady\_wind\_2ms}; primary evaluation under Strict with supplementary Grace sensitivity analysis &
Scenario-wise final error or success under task completion semantics, together with failure-mode interpretation of residual unsuccessful trials &
Cross-environment filtering of the controller architecture under autopilot-mediated dynamics, delay-sensitive settling, and stronger transit-to-hold coupling \\

Hardware Stage~A &
Vicon-tracked Tello platform &
Repeated single-waypoint trials and short multi-waypoint runs using a compact baseline, an aggressive fast-approach baseline, and the bare layered controller &
Reach rate, reach time, peak overshoot, terminal position error, terminal yaw error, and short-window hold motion &
Physical-platform method discrimination: test whether the principal structural differences identified in Phase~I remain visible on hardware \\

Hardware Stage~B &
Vicon-tracked Tello platform &
Repeated end-to-end multi-waypoint mission using only the final controller family selected after Phase~II &
Mission completion, total mission time, per-waypoint terminal position or yaw error, mean overshoot, and hold-quality metric &
Final hardware validation of the selected controller family under the finalized real-platform protocol \\
\bottomrule
\end{tabularx}
\end{table*}

\begin{table}[!t]
\centering
\caption{Evaluation semantics used in the Phase~II sensitivity analysis. Strict is the primary reporting reference in Phase~II, whereas Grace is reported only as a supplementary task-semantic sensitivity analysis. Thresholds and dwell windows are fixed by the evaluation protocol.}
\label{tab:strict_grace}
\scriptsize
\setlength{\tabcolsep}{3pt}
\renewcommand{\arraystretch}{1.05}
\begin{tabularx}{\columnwidth}{
>{\raggedright\arraybackslash}p{1.15cm}
>{\raggedright\arraybackslash}p{2.45cm}
>{\raggedright\arraybackslash}p{2.0cm}
>{\raggedright\arraybackslash}X}
\toprule
\textbf{Crit.} & \textbf{State Bounds} & \textbf{Temporal Condition} & \textbf{Interpretation} \\
\midrule
Strict &
\shortstack[l]{$e_p(t)\leq \epsilon_p$\\ $e_\psi(t)\leq \epsilon_\psi$\\ $s(t)\leq \epsilon_v$} &
Must hold for all $t\in[T_1,T_2]$ &
Hard stationary completion; primary reporting reference in Phase~II \\

Grace &
\shortstack[l]{$e_p(t)\leq R_p$\\ $e_\psi(t)\leq R_\psi$} &
Must stay inside the tube for at least $T_d$ over $[T_g,T_g+T_d]$ &
Task-semantic completion; supplementary sensitivity check \\
\bottomrule
\end{tabularx}
\end{table}

\section{Results}

Results are organized by evaluation phase rather than by implementation chronology. Phase~I narrows the design space through benchmark comparison in PyBullet, with emphasis on what separates average-case waypoint pursuit from disturbance-robust terminal regulation. Phase~II then filters those structural choices in PX4/Gazebo through a small set of representative architectural checkpoints rather than an exhaustive version log, and it identifies the final controller family to be carried forward. The final hardware subsection mirrors this same logic in physical experiments: Stage~A reports a compact hardware comparison for method discrimination, whereas Stage~B reports the final hardware validation of the selected controller family.

\subsection{Phase I Comparisons in the PyBullet Benchmark}

Phase~I serves as the early comparative benchmark stage of the paper. Its role is not merely to identify the highest-scoring controller in PyBullet, but to determine which controller properties are actually responsible for robust terminal behaviour under repeatable wind perturbations. The main outcome of this phase is therefore methodological: compact optimal-style and heuristic baselines can already generate acceptable average-case approach trajectories, which means that smooth transit alone is not the central difficulty of the task. The decisive differences emerge after nominal arrival, where wind persistence and late-stage settling expose whether a controller can regulate the target region rather than merely pass through it cleanly.

Under this interpretation, Table~\ref{tab:phase1_quantitative} functions as a structural filter rather than as a pure leaderboard or a final statistical robustness claim. Classical and aggressive baselines remain competitive in coarse waypoint pursuit, but the bare layered controller is the strongest benchmark performer that does not rely on target-specific priors. Under the wind protocol, it achieves a late-stage position mean of 0.024~m and a late-stage position standard deviation of 0.017~m across 10 trials, while maintaining 0.0023~m mean and 0.0013~m standard deviation in the no-wind evaluation. These results are important because they show that the layered controller is improving the terminal regime itself, not simply making the nominal approach look smoother. This is the point at which the core architectural decisions of Section~\ref{sec:method} become justified by evidence: approach generation and terminal regulation should be separated, transit-dominant action should fade out near the target, and near-target behaviour should be governed by explicit mode-dependent supervision.

An optional target-conditioned refinement pushes the benchmark score further, reducing the wind-condition late-stage errors to 0.019~m mean and 0.013~m standard deviation under the same evaluation. However, this refinement is intentionally not treated as the main method. Its value is diagnostic rather than architectural: it quantifies how much additional benchmark performance can be extracted when fixed target-distribution information is allowed, but it does not change the central Phase~I conclusion. For the remainder of the paper, Phase~I therefore carries forward only a small and structurally interpretable subset: a compact baseline, an aggressive baseline, and the bare layered controller as the main reference set for Phase~II. The benchmark-specific refinement is retained only as an upper-bound reference and not as a method-level candidate for cross-environment or hardware validation.

\begin{figure*}[!t]
\centering
\includegraphics[width=\textwidth]{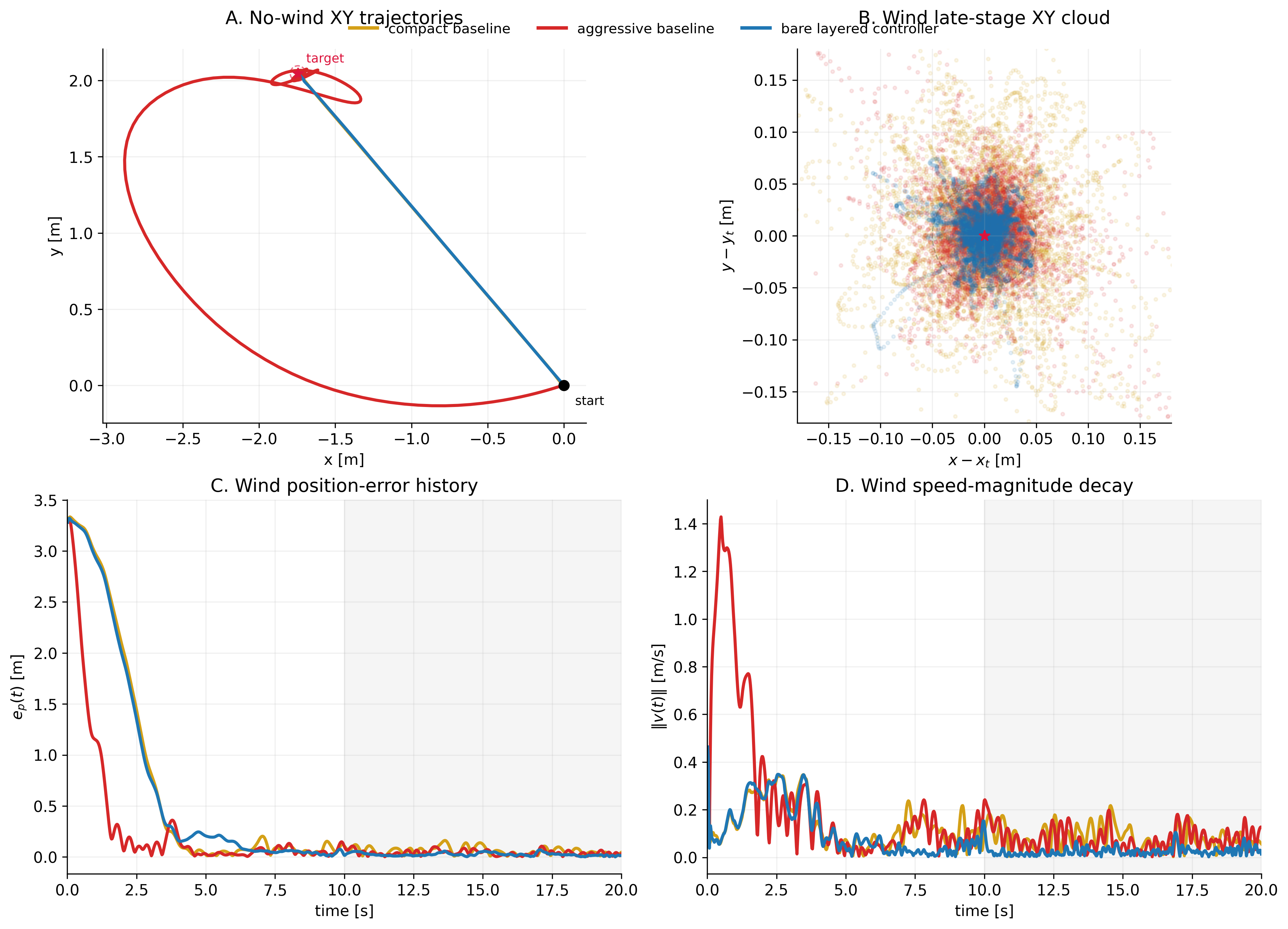}
\caption{Phase-I comparison in the PyBullet benchmark. Under no wind, all three representative controllers generate nominally acceptable approach trajectories. Under stochastic wind, however, the late-stage cloud, residual position-error history, and speed-decay traces reveal a clear separation in terminal behaviour. The bare layered controller produces the tightest late-stage dispersion and the cleanest post-arrival regulation, supporting its role as the strongest non-target-specific controller carried forward from the benchmark stage.}
\label{fig:phase1_comparison}
\end{figure*}

\begin{table*}[!t]
\centering
\caption{Quantitative Phase-I comparison in the PyBullet benchmark environment. The benchmark-specialized target-bias extension achieves the best benchmark score, but the bare layered controller is treated as the main method carried forward because it is the strongest non-target-specific controller in the Phase-I study.}
\label{tab:phase1_quantitative}
\scriptsize
\setlength{\tabcolsep}{7pt}
\renewcommand{\arraystretch}{1.12}
\begin{tabular}{p{4.1cm}cccc}
\toprule
\textbf{Controller} & \textbf{Wind Mean} & \textbf{Wind Std.} & \textbf{No-Wind Mean} & \textbf{No-Wind Std.} \\
\midrule
\texttt{controller\_decoy.py} & 0.06924 & 0.04473 & 0.00099 & 0.00096 \\
\texttt{controller\_group.py} & 0.04429 & 0.03323 & 0.00474 & 0.00169 \\
bare \texttt{controller.py} & 0.02369 & 0.01713 & 0.00229 & 0.00132 \\
\shortstack[l]{\texttt{controller.py} +\\ \texttt{target\_bias\_map.npz}} & 0.01984 & 0.01297 & 0.00229 & 0.00132 \\
\bottomrule
\end{tabular}
\end{table*}

\subsection{Phase II Cross-Environment Closed-Loop Evaluation}

Phase~II is the cross-environment filtering stage of the paper. Whereas Phase~I identifies a promising controller architecture in the PyBullet benchmark, Phase~II asks whether that architecture remains defensible once the evaluation loop becomes more physically demanding and more sensitive to coupling, delay, and terminal-settling semantics. The phase therefore begins with an intentionally important negative-transfer result: the most direct migration of the benchmark ancestor does not survive the harder PX4/Gazebo loop. When benchmark-specific priors are carried over unchanged, the controller collapses on the representative hard cases, with final errors of 2.4326~m on \texttt{combo\_hard}, 2.2792~m on \texttt{gust\_step\_y}, and 2.0260~m on \texttt{steady\_wind\_2ms}. This failure is not incidental; it is the clearest evidence in the paper that strong benchmark behaviour and a controller design that remains effective in a more demanding closed loop are not the same thing.

Once those non-transferable elements are removed, the controller family re-enters a meaningful design regime. The rebuilt transferred baseline reduces final error to 0.093~m on \texttt{combo\_hard} and 0.167~m on \texttt{gust\_step\_y}, showing that the outer-loop architecture itself remains viable after simulator-specific priors are stripped away. Subsequent variants are best interpreted as structural probes rather than isolated final answers: the state-stack stage improves gust recovery, while the combo-hardened and continuous terminal-regulator branches expose different mixed-disturbance and gust-side trade-offs. These checkpoints therefore define the Phase-II trade-off surface rather than a single pointwise winner.

The final controller is therefore selected as the integrated mainline under the transfer-oriented rule stated in the experimental protocol. Under the primary Strict reporting reference, the selected controller remains competitive on both \texttt{combo\_hard} and \texttt{gust\_step\_y}; under the supplementary Grace sensitivity analysis, it reaches 0.75 success on both cases, with final errors of 0.152~m and 0.098~m, respectively. Reporting both criteria separates structural evidence from hardware-deployment selection: branch-local endpoint error characterizes individual design probes, whereas the carried-forward mainline is determined by the transfer rule.

Residual unsuccessful trials are therefore interpreted as terminal-quality failures such as delayed settling, target re-separation after apparent arrival, or inability to satisfy the stricter completion semantics within the evaluation window, rather than as basic waypoint-pursuit failures.

\begin{figure*}[!t]
\centering
\includegraphics[width=\textwidth]{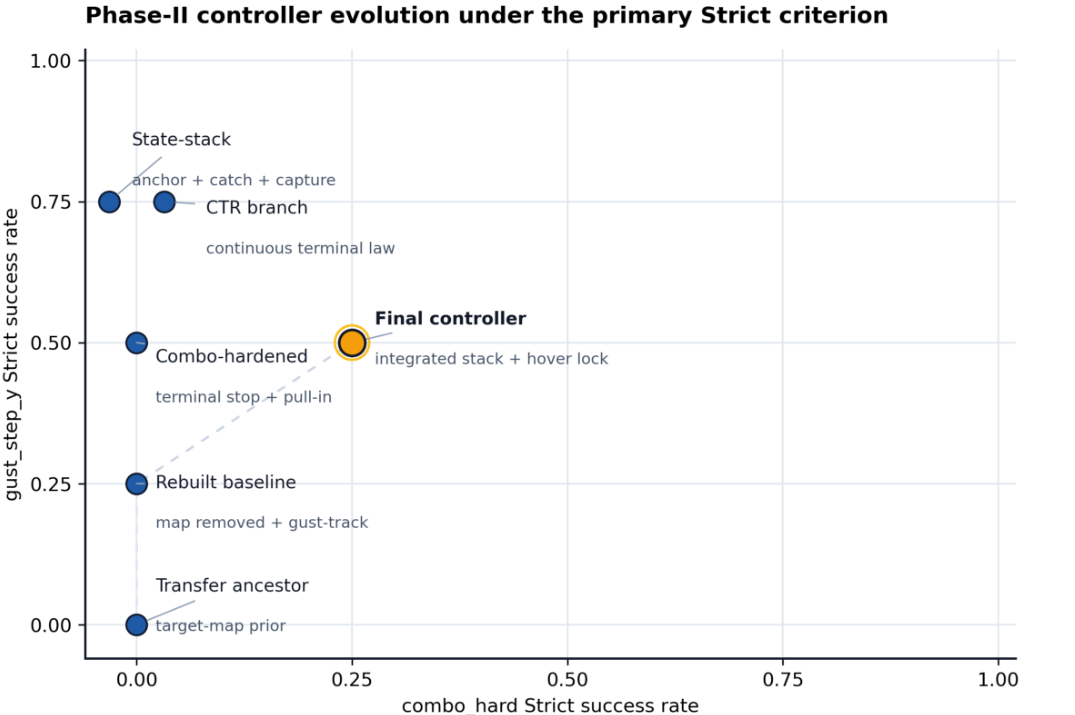}
\caption{Phase-II evolution in PX4/Gazebo. The main line moves from negative-transfer failure toward a transfer-oriented controller satisfying the carry-forward rule, while branch-local gains are treated as structural probes.}
\label{fig:phase2_tradeoff}
\end{figure*}

\begin{table*}[!t]
\centering
\begin{threeparttable}
\caption{Representative Phase-II controller versions in the PX4/Gazebo closed-loop evaluation. The table is intended as a structural filter rather than a full development log. Each metric cell reports \emph{success rate / final position error [m]}. Strict is the primary reporting reference; Grace is supplementary.}
\label{tab:phase2_versions}
\scriptsize
\setlength{\tabcolsep}{4pt}
\begin{tabular}{p{2.2cm} p{5.15cm} c c c c}
\toprule
\multirow{2}{*}{\textbf{Controller}} & \multirow{2}{*}{\textbf{Main Role}} & \multicolumn{2}{c}{\textbf{Strict}} & \multicolumn{2}{c}{\textbf{Grace}} \\
\cmidrule(lr){3-4}\cmidrule(lr){5-6}
 & & \texttt{combo\_hard} & \texttt{gust\_step\_y} & \texttt{combo\_hard} & \texttt{gust\_step\_y} \\
\midrule
\textbf{Benchmark-Transfer Ancestor}
& Direct transfer of the benchmark-era controller with target-conditioned prior still active; representative negative-transfer failure.
& 0.00 / 2.433
& 0.00 / 2.279
& --
& -- \\

\textbf{Rebuilt PX4 Baseline}
& No-map rebuilt PX4 baseline after removing the non-transferable target prior; confirms that the core feedback loop remains viable.
& 0.00 / 0.093
& 0.25 / 0.167
& 0.00 / 0.154
& 0.25 / 0.182 \\

\textbf{Layered State-Stack}
& Near-target state-stack stage; anchor, catch, and capture semantics make gust recovery structurally reproducible rather than probe-specific.
& 0.00 / 0.087
& 0.75 / 0.089
& --
& -- \\

\textbf{Combo-Hardened Branch}
& Mixed-disturbance hardening branch; pushes \texttt{combo\_hard} terminal error down aggressively, at the cost of a stiffer gust-side trade-off under Strict.
& 0.00 / 0.075
& 0.50 / 0.196
& 0.75 / 0.159
& 1.00 / 0.057 \\

\textbf{Continuous Terminal Regulator}
& Continuous terminal-control branch; remains highly competitive on gust recovery and serves as the strongest non-selected counterexample.
& 0.00 / 0.078
& 0.75 / 0.086
& 0.625 / 0.228
& 0.875 / 0.067 \\

\textbf{Balanced Final Controller}
& Final selected controller family; retained for hardware deployment because it satisfies the transfer-oriented carry-forward rule.
& 0.25 / 0.294
& 0.50 / 0.166
& 0.75 / 0.152
& 0.75 / 0.098 \\
\bottomrule
\end{tabular}
\begin{tablenotes}[flushleft]
\footnotesize
\item Missing entries (`--') indicate that no directly comparable Grace-summary artifact was retained in the repository for that representative checkpoint.
\item The final controller row is retained because it satisfies the transfer-oriented carry-forward rule; branch-local gains are interpreted as structural evidence rather than as hardware-selection criteria.
\end{tablenotes}
\end{threeparttable}
\end{table*}

\subsection{Real-Platform Results}
\label{sec:hardware_results}

This subsection reports the completed real-platform evaluation in two blocks that mirror the logic of the first two phases. All hardware experiments are executed on the Vicon-tracked Tello platform under fixed command bounds, common waypoint definitions, and controlled terminal thresholds. PX4/Gazebo is used earlier as a harder closed-loop filter rather than as a Tello-specific simulator. Stage~A corresponds to Phase~I and asks whether the principal controller distinctions observed during the early benchmark comparison remain visible on the physical platform. Stage~B corresponds to Phase~II and asks what the selected final controller family can achieve once the cross-environment selection process is complete.

In the Stage-A hardware comparison, a compact classical baseline, an aggressive fast-approach baseline, and the bare layered controller are evaluated over 10 repeated single-waypoint trials, with 3 short multi-waypoint runs used as consistency checks. The comparison shows that the layered architecture bridges the gap between fast transit and terminal precision on physical hardware. Specifically, the aggressive baseline exhibits the fastest reach time but suffers from severe overshoot and rebound, the compact baseline avoids overshoot but is overly slow and struggles to reject residual drift, and the layered controller achieves competitive transit times while maintaining the tightest terminal error and stable hold. Taken together, under the present platform and protocol, these results indicate that controllers without layered terminal regulation tend to exhibit poorer terminal capture, larger overshoot, and weaker short-window hold quality even when gross waypoint pursuit remains feasible. Table~\ref{tab:hardware_results} summarizes reach rate, reach time, terminal position and yaw error, and terminal-behaviour indicators. This block answers whether the structural ranking identified in Phase~I remains meaningful once the controller leaves simulation.

In the Stage-B hardware validation, only the final controller family selected in Phase~II is carried forward. This experiment evaluates end-to-end multi-waypoint performance over 5 repeated runs of a 6-waypoint mission with tighter terminal settings and richer logging. The selected controller achieves consistent mission completion (30/30 reached segments at the 0.5~m tolerance), with mean total mission time 34.2~s, mean per-waypoint position error 0.46~m, mean per-waypoint yaw error 0.12~rad, mean overshoot 0.25~m, and hold-quality metric 0.08~m/s RMS. The dominant remaining limitation is the inability to reliably tighten the residual error below 0.3~m, observed primarily under the transition to hover where unmodeled Tello aerodynamic drag and command delay dominate the outer-loop bandwidth. Taken together, the two hardware blocks close the paper narrative on the chosen physical platform in the same order as the earlier phases: Phase~I narrows the candidate set, Phase~II selects the final controller family, and the real-platform study validates those two decisions on Tello hardware.

\begin{figure*}[!t]
\centering
\includegraphics[width=0.97\textwidth,height=0.68\textheight,keepaspectratio]{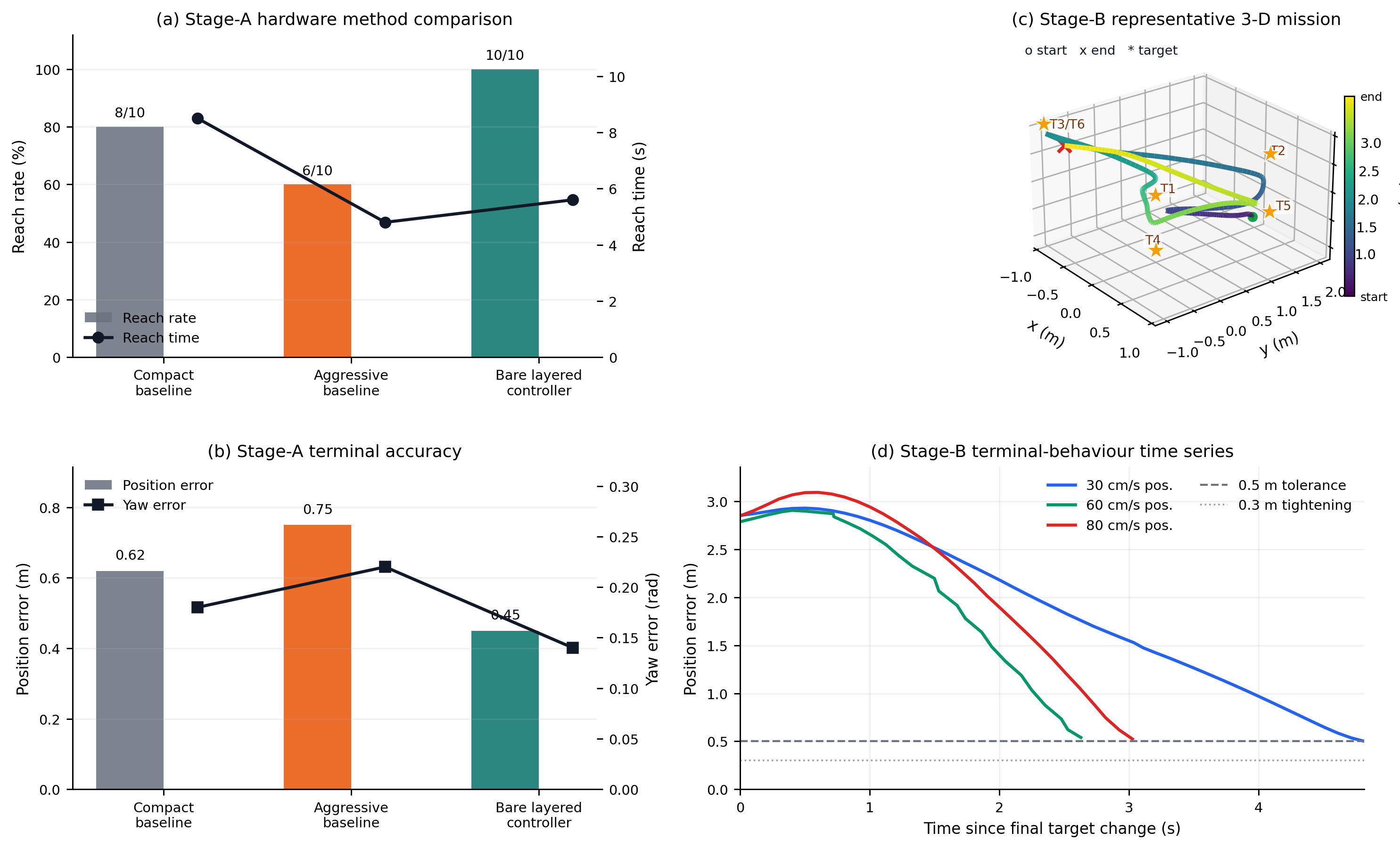}
\caption{Real-platform results across the two hardware studies. Stage~A compares representative controllers for method discrimination; Stage~B reports final-controller validation through mission trajectories, quantitative spread, and terminal-behaviour time series.}
\label{fig:hardware_results}
\end{figure*}

\begin{table*}[!t]
\centering
\caption{Hardware results summary for Stage-A method comparison and Stage-B final-controller validation.}
\label{tab:hardware_results}
\begin{threeparttable}
\scriptsize
\setlength{\tabcolsep}{3pt}
\renewcommand{\arraystretch}{1.1}
\begin{tabularx}{\textwidth}{>{\raggedright\arraybackslash}p{2.35cm} >{\centering\arraybackslash}p{1.45cm} >{\raggedright\arraybackslash}p{2.25cm} >{\centering\arraybackslash}p{1.2cm} >{\centering\arraybackslash}p{1.25cm} >{\centering\arraybackslash}p{1.15cm} >{\raggedright\arraybackslash}p{1.55cm} >{\raggedright\arraybackslash}p{1.45cm} >{\raggedright\arraybackslash}X}
\toprule
\textbf{System / Controller} & \textbf{Trials} & \textbf{Completion / Reach} & \textbf{Time} & \textbf{Pos. Err.} & \textbf{Yaw Err.} & \textbf{Overshoot / Rebound} & \textbf{Hold Quality} & \textbf{Notes} \\
\midrule
\multicolumn{9}{l}{\textbf{Part A: Stage-A Hardware Method Comparison}} \\
Compact classical baseline & 10 SW + 3 MW checks & 8/10 at 0.5~m & 8.5~s & 0.62~m & 0.18~rad & Low overshoot & Weak drift rejection & Slow approach; residual drift remains visible \\
Aggressive fast-approach baseline & 10 SW + 3 MW checks & 6/10 at 0.5~m & 4.8~s & 0.75~m & 0.22~rad & Severe rebound & Weak hold & Fast transit but poor terminal capture \\
Bare layered controller & 10 SW + 3 MW checks & 10/10 at 0.5~m & 5.6~s & 0.45~m & 0.14~rad & Bounded rebound & Stable hold & Best terminal-error trade-off among Stage-A controllers \\
\midrule
\multicolumn{9}{l}{\textbf{Part B: Stage-B Final-Controller Validation}} \\
Balanced Final Controller & 5 missions & 30/30 segments at 0.5~m & 34.2~s mission & 0.46~m mean/waypoint & 0.12~rad mean/waypoint & 0.25~m mean & 0.08~m/s RMS & Consistent 0.5~m capture; 0.3~m tightening remains incomplete \\
\bottomrule
\end{tabularx}
\begin{tablenotes}[flushleft]
\footnotesize
\item SW denotes single-waypoint trials and MW denotes short multi-waypoint consistency checks; Stage-A quantitative reach rates are computed from the 10 repeated SW trials.
\item Stage-B values summarize 5 repeated runs of a 6-waypoint mission under the final layered controller, giving 30 reached segments at the 0.5~m tolerance.
\item Segment reach is counted using the stated protocol tolerance, while position and yaw errors are terminal per-waypoint summary values over the evaluated segments.
\item Stage-A overshoot and hold entries are qualitative labels from the repeated trial logs; Stage-B hold quality is the RMS residual speed during the final hold window.
\end{tablenotes}
\end{threeparttable}
\end{table*}

\section{Discussion and Limitations}

The main strength of this paper is not that it introduces a previously unseen control component, but that it organizes known ideas in UAV control into a structured and testable terminal-regulation architecture~\cite{Mellinger2011,Cabecinhas2014,Chen2016,Liberzon2003}. Disturbance observers, integral action, phase-dependent supervision, and smooth approach shaping are established tools; the contribution here is the way they are combined and filtered through staged evidence. This framing also explains why compact baselines remain competitive under average-case waypoint metrics while the layered controller is most useful when the evaluation emphasizes late-stage settling, rebound suppression, disturbance-robust hold, and behaviour under harder closed-loop conditions.

The same structure has an engineering cost. The controller is designed around a bounded outer-loop interface, persistent internal state, disturbance-aware handoff, and supervisory terminal behaviour, which make deployment through practical velocity-command stacks plausible. However, it remains parameterized and mode-rich: interacting gains, thresholds, and transition conditions make attribution harder and retuning more expensive when the platform, sensing quality, or disturbance profile changes. The architecture is therefore best viewed as a hardware-validated research controller rather than an industrially certified flight stack. Its outer-loop organization is portable to UAV stacks exposing bounded velocity commands and reliable state feedback, but the parameterization does not transfer automatically across PyBullet, PX4/Gazebo, and Tello hardware.

Several limitations remain. The method is an engineered supervisory controller without a formal closed-loop proof of stability or recursive feasibility, and it assumes accurate state feedback and velocity-level command authority rather than direct thrust or attitude control. The real-platform experiments establish that the proposed structure can be deployed and evaluated on a Vicon-tracked Tello stack, but they do not by themselves prove broad hardware generality. The evaluation still covers one benchmark family, a limited set of PX4/Gazebo disturbance scenarios, one indoor motion-capture hardware setting, and short-duration waypoint missions. The Strict/Grace sensitivity analysis strengthens the interpretation of terminal behaviour, but also shows that some conclusions depend on the operational definition of completion. Broader robustness across vehicle classes, onboard-only state estimation, outdoor wind fields, and long-duration deployment remains future work.

\section{Conclusion}

This letter presented a layered terminal-control architecture and staged cross-environment evaluation for disturbance-robust multi-waypoint UAV regulation. Phase~I used a PyBullet benchmark to identify controller structure associated with late-stage wind robustness, while Phase~II carried only the transferable controller core into a harder PX4/Gazebo closed loop and selected the final controller family using the predeclared transfer-oriented rule rather than a single-scenario endpoint score. The hardware study then closes this evidence chain on one Vicon-tracked Tello platform. Overall, the paper contributes both a controller architecture and an evaluation logic: benchmark performance becomes more informative when the main controller structure is analyzed separately from benchmark-specific refinements and then tested under progressively harder closed-loop conditions.

\end{document}